\documentclass[conference,peerreview]{IEEEtran}

\usepackage{booktabs, multicol, multirow}
\usepackage[acronym,nomain]{glossaries}
\usepackage[inline]{enumitem}
\usepackage{algorithm}
\usepackage{algorithmicx}
\usepackage{algpseudocode}
\usepackage{amsmath}
\usepackage{amsfonts}
\usepackage{subfigure}
\usepackage{grffile}
\usepackage{tabularx}
\usepackage{colortbl}
\usepackage{hhline}
\usepackage{hyperref}
\usepackage{here}
\usepackage{array}
\usepackage{rotating}
\usepackage{amssymb}
\usepackage{flushend}

\providecommand{\tabularnewline}{\\}

\usepackage[colorinlistoftodos,prependcaption,textsize=tiny]{todonotes}

\newboolean{showcomments}
\setboolean{showcomments}{false} 
\ifthenelse{\boolean{showcomments}}
{\newcommand{\nb}[2]{
  \fcolorbox{black}{yellow}{\bfseries\sffamily\scriptsize#1}
  {\sf\small$\blacktriangleright$\textit{#2}$\blacktriangleleft$}
 }
 
}
{\newcommand{\nb}[2]{}
 
}

\newcommand\martin[1]{\nb{Martin}{#1}}

\usepackage{pgf}
\usepackage{tikz}
\usetikzlibrary{arrows,automata}
\usetikzlibrary{shapes.misc}

\tikzset{cross/.style={cross out, draw=black, minimum size=3*(#1-\pgflinewidth), inner sep=0pt, outer sep=0pt},
	cross/.default={10pt}}

\usepackage{array}
\usepackage{url}

\def\BibTeX{{\rm B\kern-.05em{\sc i\kern-.025em b}\kern-.08em
    T\kern-.1667em\lower.7ex\hbox{E}\kern-.125emX}}
\begin{document}

\title{DataOps for Societal Intelligence: a Data Pipeline for Labor Market Skills Extraction and Matching
}
\author{\IEEEauthorblockN{Damian A. Tamburri}
\IEEEauthorblockA{\textit{Technical University Eindhoven} \\
\textit{Jheronimus Academy of Data Science}\\
s'Hertogenbosch, The Netherlands \\
d.a.tamburri@tue.nl}
\and
\IEEEauthorblockN{Willem-Jan Van Den Heuvel}
\IEEEauthorblockA{\textit{Tilburg University} \\
\textit{Jheronimus Academy of Data Science}\\
s'Hertogenbosch, The Netherlands \\
W.J.A.M.v.d.Heuvel@jads.nl}
\and
\IEEEauthorblockN{Martin Garriga}
\IEEEauthorblockA{\textit{Tilburg University} \\
\textit{Jheronimus Academy of Data Science (JADS)}\\
s'Hertogenbosch, The Netherlands \\
m.garriga@uvt.nl}
}

\maketitle

\begin{abstract}


Big Data analytics supported by AI algorithms can support skills localization and retrieval in the context of a labor market intelligence problem. We formulate and solve this problem through specific DataOps models, blending  data sources from administrative and technical partners in several countries into cooperation, creating shared knowledge to support policy and decision-making. 
We then focus on the critical task of skills extraction from resumes and vacancies featuring state-of-the-art machine learning models. 
We showcase preliminary results with applied machine learning on real data from the employment agencies of the Netherlands and the Flemish region in Belgium. The final goal is to match these skills to standard ontologies of skills, jobs and occupations.

\end{abstract}


\ifCLASSOPTIONpeerreview
 \begin{center} \bfseries EDICS Category: 3-BBND \end{center}
 \fi
%
\IEEEpeerreviewmaketitle

\section{Introduction}\label{sec:intro}
One of the major downsides of our digital economy is the increasing predominance with which technology has substituted human labour and skills. On one hand, computing skills are increasing in demand and low in supply. On the other hand, formerly needed skills and labour areas often become subject to crisis. 
We argue that these are \emph{*societal*} shortcomings and reflect just one example of the relative lack of \emph{societal intelligence}; that is, the ability of our society at large to harness the digital ecosystem and turn it into an opportunity. In particular, Big Data analytics supported by DataOps methods and AI algorithms now offer the potential to make the act of finding the right skills into a labor market intelligence problem. The challenges we face in this context are manifold:

(i) Data Integration of data sources from administrative and technical partners (job seekers, employers, trainers, public authorities, etc) in several countries into cooperation, creating shared knowledge to support policy and decision-making;

(ii)  Extracting available/required skills and employment options, specially in low-compliance areas (e.g., borderlands); and,

(iii) Matching skills from resumes with vacancies automatically, leveraging structured information about Skills, Competences and Opccupations from ontologies and taxonomies such as the EU standards ESCO\footnote{https://ec.europa.eu/esco/} 
and ISCO~\cite{hoffmann2003international}. 

(iv) Assess, Improve and Extend the existing ontologies and taxonomies with the descriptions of novel, arising job profiles and skills. 

In this context, the main contributions of this paper target the following research questions:

\noindent \textbf{RQ1.} \textit{How to model a DataOps Pipeline to enable Societal and Labour Market Intelligence in the context of skills extraction and matching? }

We design such pipeline addressing the aforementioned challenges, 
to move from Big Data to Knowledge. Then we illustrate 
how to automate and scale the analysis of job vacancies and resumes, and then matching against representative ontologies (such as ESCO/ISCO).


\noindent \textbf{RQ2.} \textit{How to extract and match the skills (from vacancies and resumes) in an effective and efficient way? }

We instantiate a pipeline with hundreds of thousands vacancies stemming from the Dutch-Flemish labour market and then solve the skills extraction problem  featuring state-of-the-art machine learning models. 

The rest of this paper is structured as follows. Section~\ref{sota} presents related work and background definitions. Section~\ref{design} describes the DataOps design for Societal and Labour Market Intelligence.
Section~\ref{sub:big-data-solution} showcases how to apply the DataOps pipeline for the skills extraction and matching problem. Section~\ref{sec:validation} presents the validation with real data of the Dutch-Flemish cross-border labor market. Section~\ref{sec:discussion} discusses main findings and limitations. Finally, Section~\ref{conc} concludes the paper.

\section{State of the Art}\label{sota}
\subsection{Societal and Labor Market Intelligence}

Societal intelligence is the family of approaches, services, and platforms that adopt Big data analytics, machine-learning or other technologies specific to business intelligence over social, organisational, and societal data for the purpose of a more instrumented, self-sustainable society. Furthermore, Labor market intelligence (LMI) refers to the design and use of Big Data and AI algorithms and frameworks to analyze labor market information for supporting policy and decision-making~\cite{mercorio18journal,mezzanzanica2019big,uk2015lmi}. Labor market information encompasses skills, competencies, qualifications, and occupations, including the ICT techniques and services to manage such information, in particular, mobility-related services. Key to LMI is the identification and adoption of standard taxonomies for occupations and skills, to foster the circulation of information in a multi-language job market like the European one. 

\subsection{Labor Market Ontologies}


The standard ontology to classify occupations for the international labour market is ISCO~\cite{isco12ilo}~\cite{hoffmann2003international}, a four-level hierarchy; the
interested reader may consider~\cite{isco12ilo} for more details. ISCO is one of the most
adopted taxonomies in Europe and it is a reference worldwide, while the United States developed its own, namely O*NET\footnote{https://www.onetonline.org/}. 

For skills description one can find broad regional classifications such as ESCO\footnote{https://ec.europa.eu/esco/resources/data/static/model/html/model.xhtml}. ESCO is a multilingual taxonomy of European Skills, Competences, Qualifications and Occupations that extends ISCO
through a further level of fine-grained occupation descriptions, and a taxonomy of
skills, competencies and qualifications. One can navigate the ISCO-ESCO cross-linkage to
access the list of occupation examples containing a set of skills/competences and qualifications requested by the corresponding occupation.

Specific classifications can be found for each country or cross-border region. As a relevant example within this work, think of Competent\footnote{https://competent.vdab.be/competent/}, developed in Belgium and recently extended to the Netherlands as Competent-NL\footnote{http://production.competent.be/competent-nl/main.html}. 
Competent and other fine-grained classifications can be easily linked to ISCO and ESCO through mapping tables~\cite{esco16mapping}.

\subsection{DataOps}


Like DevOps, DataOps aims to combine the production, operation and delivery (of data) into a single, agile practice that directly supports specific business functions to improve quality, speed, and collaboration and promote a culture of continuous improvement~\cite{ereth2018dataops}. A DataOps methodology combines and interconnects data engineering, data integration, data quality, and data security/privacy~\cite{palmer2015devops} to deliver data from its source to the person, system, or application that can turn it into business value~\cite{nexladataops}.

\section{Big data for  Societal Intelligence: A proof-of-concept for the Labor Market}\label{design}

\begin{figure*}[htb]
\centering
\includegraphics[width=\textwidth]{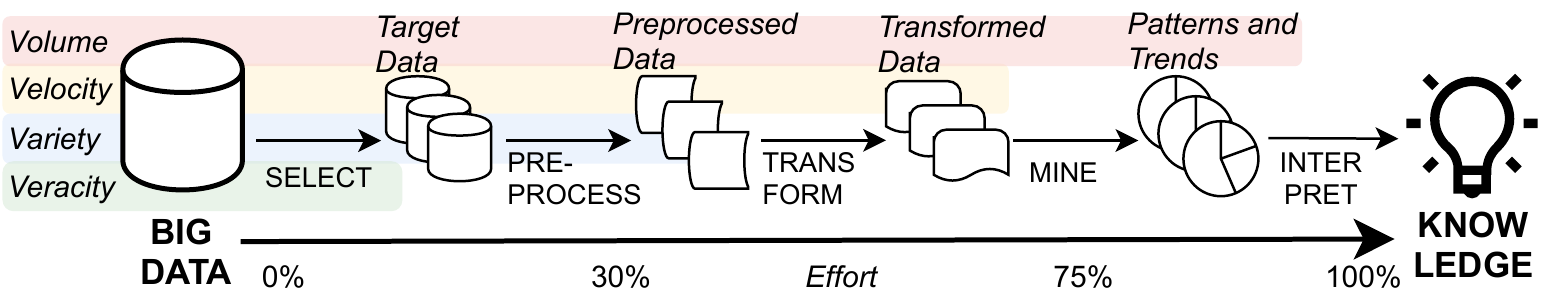}
\caption{Big Data enables Societal and Labor Market Intelligence. The process specifies which Vs of Big Data are involved in each step (adapted from~\cite{mezzanzanica2019big})}\label{fig:comm}
\end{figure*}

The extraction of knowledge from (Big) labor market data has been addressed using the knowledge discovery in databases (KDD) approach as baseline~\cite{fayyad1996kdd} -- as shown in Figure~\ref{fig:comm}, together with the span of the Vs of Big Data throughout the process. 

\paragraph{Step 1: Select Data Sources} Each Web source is evaluated and ranked to guarantee its reliability -- e.g., vacancy publication date, update frequency, (un)structured data, privacy restrictions, etc. Heterogeneous Data sources are considered here, including but not limited to, governmental employment offices (from different regions/countries), private Websites (such as indeed.com) and social networks (such as Linkedin). At the end of the process, the data sources are selected. This step deals with the 4 Vs of Big Data including \textit{veracity}, intended as the biases, noise and/or abnormalities in data.

\paragraph{Step 2 Pre-processing Selected Data} This includes data cleansing tasks to remove noise from the data or inappropriate outliers (if any), deciding how to handle missing data, as well as removing duplicated entries. Data quality and cleaning tasks are indeed \textit{mandatory} steps of any data-driven decision-making approach for guaranteeing the trust on the overall process, intended as the "the extent to which data is accepted or regarded as true, real and credible"~\cite{mezzanzanica2019big}. This step reduces the complexity of the Big Data scenario, by addressing the \textit{variety} dimension.

\paragraph{Step 3: Transformation}
This step transforms data into a unified model, which depends on the goal of the process, by means of data reduction and projection techniques. 
It usually involves ETL (Extract, Transform, Load) techniques typical of Big Data Integration scenarios~\cite{vassiliadis2002conceptual}. 
Around 70\% of the effort of the overall process is spent on Data Selection, Pre-processing and Transformation, to provide input data in the amount, structure and format that suit each Data Mining task perfectly~\cite{garcia15data}.

A crucial transformation for Societal and Labour Market Intelligence consists of extracting the important information -- i.e.,  the skills (hard and soft) in the job vacancy text. More formally, the transformation function takes a job vacancy $j_k \in J$ as input and extracts a set of skills $S_k=\{ s_1,...,s_n\}$ where $S_k \in P(S)$, the powerset of skills  $S$. Note that the skills set $S$ evolves throughout time as novel skills arise. 

\paragraph{Step 4: Data Mining and Machine Learning} This step requires to identify proper AI algorithms (e.g., classification, prediction, regression, clustering), searching for patterns of interest. In our context, it usually involves text classification algorithms to map vacancies/resumes into one of several predefined classes in a taxonomy or ontology of job profiles and skills, such as ISCO, ESCO and/or Competent.
More formally, categorization assigns a Boolean value to each pair $(j_k,c_i)\in JxC$ -- where \textit{J} are job vacancies and \textit{C} the predefined categories (of a given taxonomy) --  \textit{true} if $j_k$ belongs to $c_i$ and \textit{false} otherwise. This categorization task can be solved through machine learning: let $J = {J_1, ..., J_n}$ be a set of job vacancies and the classification of $J$ under the taxonomy consists of $|O|$ independent problems of classifying each job vacancy $J_i\in J$ under a given taxonomy occupation code $o_i$ for $i=1...|O|$. Then, a classifier is a function $\phi : JxO \xrightarrow{} \{0,1\}$.

\paragraph{Step 5: Interpretation and Feedback} Finally, this step employs visual paradigms to represent the resulting knowledge, according to the ultimate goal. In our context, this step concerns the users' ability to understand the data. For example, public administrations might be interested in identifying the most requested occupations in a cross-border area; companies might focus on monitoring skills trend and novel, arising skills and occupations, so that they can design training paths.

\section{A DataOps Processing Solution for Skills Extraction and Matching}
\label{sub:big-data-solution}

\begin{figure*}
  \centering
  \includegraphics[width=1\textwidth]{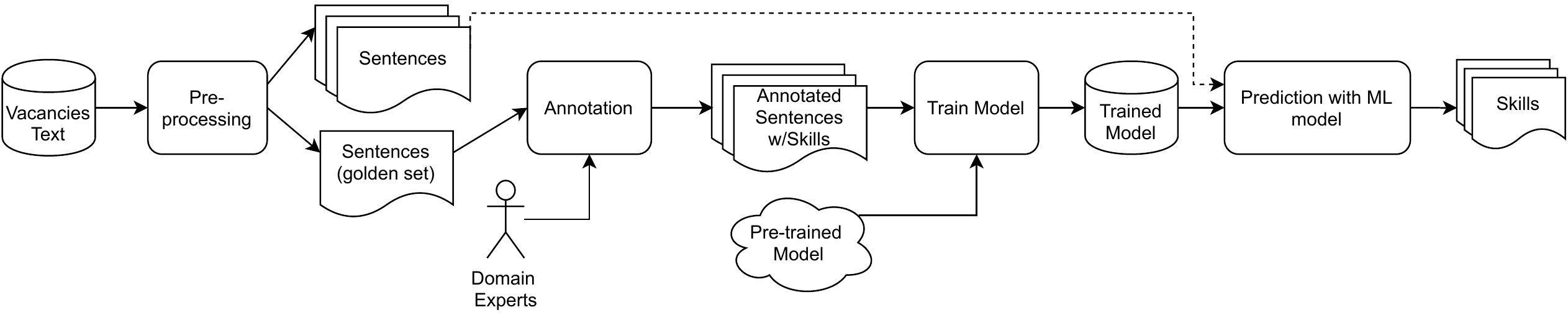}
  \caption{Model training and prediction pipelines for skills extraction and matching.}\label{fig:training-pipeline}
\end{figure*}

From the steps described in the previous section, we focused our prototype on the critical task of skills extraction and matching. This entails a DataOps pipeline combining Big Data and Machine Learning models to process text in vacancies and resumes, and then extract the sentences containing skills. Figure~\ref{fig:training-pipeline} shows the Model Training and Prediction pipelines, detailed below.

\paragraph{Data Pre-processing}

Input data must be provided in the amount, structure and format that suit each Data Mining task perfectly. Low quality data will lead to low quality results: \textit{garbage in, garbage out}~\cite{mierswa2006yale}. 
Data Pre-processing may feature several tasks, typically including Data Cleaning, Transformation, Integration, Normalization, Missing Data Imputation and Noise Identification~\cite{garcia15data}.

Given the characteristics of our input (vacancies text), several text preprocessing steps are necessary, mainly comprising (but not limited to):

\begin{itemize}
    \item Distinguish end-of-sentence periods ( . ) from others (e.g. abbreviations, acronyms, websites, etc.).
    \item Distinguish end-of-sentence line breaks (\textit{\textbackslash	n}) from spurious ones.
    \item Trim leading/trailing spacing/tabbing.
    \item Identify and replace special characters used as bullet points (such as hyphens, dashes or even 'o').
    \item Filter stopwords with little lexical meaning, such as pronouns and articles.
\end{itemize}

The output of this step is a list of processable sentences, ready to be used by the Machine Learning algorithm. 

\paragraph{Sentences Annotation for Learning}

Each sentence in the training set must be annotated by domain experts: the ones that contain a skill are annotated with a 1, otherwise with a 0.

In the case of ambiguous sentences, we created a codebook with help of the domain experts as an aid to identify and correctly annotate the sentences, with the following, previously agreed definitions:

 \textbf{Hard skill}: Is a professional skill, activity or knowledge. These are related to functions, processes and roles in organizations and are needed in practice to successfully complete a specific task. For example: 

\textit{Your work consists of checking and controlling cleaning activities of the chambermaids, opening and / or closing your service, communication with the reception and / or guests and light administrative work.
}

An example of vacancy sentences without hardskill:

 \textit{   It is an international company that provides management, advisory and engineering services on socially relevant issues related to mobility, environment \& space, buildings and water.}

 \textbf{Soft skill}: Is a personal, emotional, social or intellectual skill, also called \textit{behavioral skill} or competence. Soft skills are often seen as good-to-have in addition to the hard skills. For example:  

 \textit{   You are able to maintain customer relationships, identify opportunities and utilize and expand your network;}

\paragraph{Model Training}


First, the model is trained on unlabeled data over different pre-training tasks. This reduces the need for many heavily-engineered task-specific architectures. Pre-trained models can be fine-tuned with just one additional output layer to create state-of-the-art models for a wide range of tasks, such as question answering and language inference, without substantial task-specific architecture modifications~\cite{devlin2018bert}.

\paragraph{Supervised Training (fine-tunning)}

Then a training set is given to form a description that can be used to predict unseen examples. This training set ('train' data) is typically a set of labeled instances~\cite{garcia15data}. 
This is pre-processed (as described above), tokenized and converted into the format that the model understands -- typically in the form of csv or tsv files, containing the columns with the annotations, the vacancies/resumes and skills ids for traceability, and the actual sentence text.

At this point the model can be actually trained for the needed purposes. Early stopping with a configurable number of epochs is recommended, as the training theoretically should finish when the model achieves the optimal state under the setup conditions, avoiding overfitting.

Afterwards, the accuracy can be verified using the 'dev' validation dataset, a labeled set not directly used for training. Note that currently a small set of manually annotated data is used to train the model. Increasing the training set even by ~500-600 examples can boost predictions accuracy.

All in all, the output of this process is the trained model which can be used for further predictions. Note that, compared to pre-training, fine-tuning is relatively inexpensive. All of the results in the following section can be replicated in at most 1 hour on a single Cloud TPU, or a few hours on a GPU, starting from the exact same pre-trained model.

\paragraph{Predictions with the Pre-trained ML Model}

At this point the model is ready to make predictions, according to the pipeline depicted in Figure~\ref{fig:training-pipeline}.  Recall that the goal of the pipeline is to classify (predict) sentences describing skills from a vacancy text. 

Pre-processing is equal to the one performed for training. Then the sentences extracted from the processed vacancies/resumes text (in the proper tsv/csv format) are given as input to the model, pre-trained and fine-tuned during the training pipeline. 
The output contains the predictions for each sample,  with the original text, ids and class probabilities (whether the sentence contains a skill or not).

\section{Validation: The Case on the cross-border Dutch-Flemish labor market }\label{sec:validation}

\subsection{Context}

The case-study in question exemplifies the applicability and impact of the proposed DataOps pipeline, featuring the economy and the labor market in Dutch border regions which are malfunctioning w.r.t. other regions in the Netherlands. 
At the same time, there are opportunities on the other side of the border (i.e., Belgium, Germany) that are not or insufficiently utilized. Many factors can impede cross-border work:
differences in language and culture, institutional and administrative differences, lack of cross-border infrastructure, psychological factors, economic differences and information delays (source: CPB\footnote{Netherlands Bureau for Economic Policy Analysis -- \url{https://www.cpb.nl/en}}). 




Werkinzicht\footnote{An EU Interreg project -- \url{https://werkinzicht.eu/}}, the umbrella project for this work, is working to provide a clear picture of cross-border VL-NL labor market information for job seekers, employers, educational institutions, governments and intermediaries. Cross-border job boards are available from different organizations. For example, UWV and VDAB -- Dutch and Flemish employment agencies respectively -- publish vacancies from both sides of the border. 

To date, the baseline attained in previous projects -- e.g.,  during the ESCO skills mapping pilot~\cite{esco16mapping} -- shows that the automated transformation using mapping tables introduces a high level of noise (between 81/92\%), meaning only one/two out of every ten skills represented a correctly transferred skill. Besides, the manual effort is cumbersome: a sample of 610 different skills took, on average 60 hours~\cite{esco16mapping}. This can be certainly improved through our semi-supervised approach.

\subsection{Execution and Results}

Our prototype pipelines focus on skills extraction, as shown in Figure~\ref{fig:training-pipeline}. We used three alternatives for the underlying Machine Learning Model. First, a Simple Neural Network\footnote{Source and documentation: https://tinyurl.com/y86rerbw} as baseline. Then, two alternative deployments of the best model to date for this task: BERT (Bidirectional Encoder Representations from Transformers)~\cite{devlin2018bert}, namely: BERT in the Cloud\footnote{\url{Source and documentation at: https://tinyurl.com/ybtygt52}}, running on Google Cloud with a single TPU device with 4 TPU chips and 8 cores; and BERT Local deployment\footnote{Source code and documentation at: \url{https://tinyurl.com/ycks8sp4}} running on on-premise hardware.

The input dataset was provided by the Dutch employment agency (UWV) -- consisting of millions of vacancies from year 2014 onwards. For the training and validation  activities, the domain experts curated a total of 10K vacancies that became the \textit{gold-standard}. After pre-processing it we obtained a total of 300K sentences.

\paragraph{Annotation}
Six domain experts from UWV annotated a total of 3K sentences -- around 500 sentences per person. Each sentence is annotated with a '1' if contains a skill, and as '0' if it does not. The annotation process was aided by the Codebook\footnote{\url{https://tinyurl.com/y8k34xqw} (in dutch)} discussed in Section~\ref{design}, describing what is intended as a Skill, and how to annotate the dataset. We performed two rounds of annotations plus feedback, and also cross-checked the annotations made by the experts with the ones made independently by three researchers (authors of the present paper). We calculated the Inter-rater reliability assessment for the annotation process: the annotations by the two groups (i.e., the domain experts and the researchers) were then compared in order to appraise each others annotations and comments, and to reach unanimity on the skills definition and identification. The final result for the widely known K$\alpha$~\cite{krippendorff2018content} coefficient that measures the agreement amounted to 0.86, higher than the typical reference score of 0.80 (i.e., 80\% agreement).


\paragraph{Model Training}

We trained the models according to the steps discussed in Section~\ref{sub:big-data-solution}. The baseline pre-trained model used both for BERT Cloud and BERT Local options is the BERT Multilingual Cased Model\footnote{https://github.com/google-research/bert/blob/master/multilingual.md}. From the 3K annotated sentences we adopted a 75/25 splitting for training (\textit{train} set) and validation (\textit{dev} set) respectively. 


\paragraph{Results}

\begin{figure}[htb]
\centering

\includegraphics[scale=0.4]{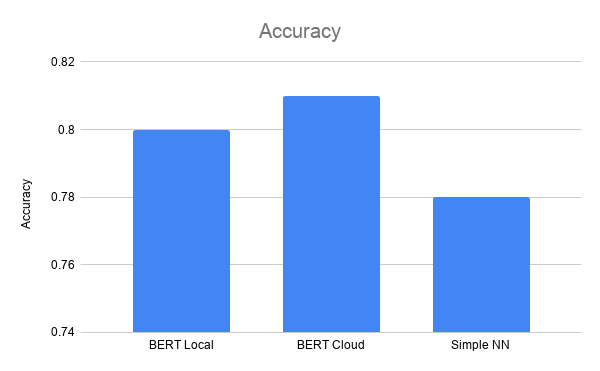}
\caption{Accuracy results (more is better) after two-steps training for the three evaluated models: BERT Cloud, BERT Local and Simple-NN}\label{fig:accuracy}

\end{figure}

The accuracy results are shown in Figure~\ref{fig:accuracy}. The best performing model in this case is BERT Cloud, with an accuracy of 0.816 after 210 training steps (epochs), followed by BERT Local (0.8) and Simple-NN (0.78). All three models show high and competitive accuracy, although it could be improved by means of training with more annotated sentences, fine-tunning hyperparameters, etc., as future work.

Finally, Table~\ref{tab:confusion-matrix} shows the confusion matrix for the best performing model -- i.e., BERT Cloud. 
Note the high recall (0.94) of annotated skills correctly predicted. Moreover, as stated by the experts from UWV, the worst case scenario are false negatives, i.e., skills that are incorrectly classified as \textit{not} skills. The model has a very low false negative rate of 0.05.


\begin{table}

\caption{Confusion Matrix for Prediction Results using BERT Cloud}
\label{tab:confusion-matrix}
\begin{center}
    
\begin{tabular}{clcc}
\toprule 
 &  & \multicolumn{2}{c}{Annotation (Actual Class)}\tabularnewline
 &  & Skill & Not Skill\tabularnewline
\midrule
\multirow{4}{*}{\begin{turn}{90}
Prediction
\end{turn}} & Predicted Skill & 0.85 (TP) & 0.30 (FP)\tabularnewline
 & Predicted Not Skill & 0.05 (FN) & 0.79 (TN)\tabularnewline
 & Accuracy & \multicolumn{2}{c}{0.816}\tabularnewline
 & Recall & \multicolumn{2}{c}{0.945}\tabularnewline
\bottomrule
\end{tabular}
\end{center}

\end{table}

\section{Discussion and Limitations}
\label{sec:discussion}


\noindent \textbf{RQ1.} \textit{How to model a DataOps Pipeline to enable Societal and Labour Market Intelligence in the context of skills extraction and matching? }

\martin{Change description to match 2 rqs}

To answer \textbf{RQ1} on the DataOps pipeline for Labour Market Intelligence, Section~\ref{design} presented our design to move from from big data to knowledge, providing societal and labour administrators with Big Data analytics and business intelligence. This allows employment agencies to have evidence of the competitive advantage that the automated analysis of Web job vacancy enables with respect to e.g., the classical survey-based analyses, in terms of (i) near-real-time labor market information; (ii) reduced time-to-market (iii) fine-grained territorial analysis, from countries down to municipalities~\cite{mezzanzanica2019big}.


To answer \textbf{RQ2} regarding the job vacancies and resumes analysis and matching, Section~\ref{analytic} illustrated shown how to scale the identification of the most requested skills and then matching against the most representative ontologies such as ISCO/ESCO and cross-border ones as Competent.

\noindent \textbf{RQ2.} \textit{How to extract and match the skills (from vacancies and resumes) in an effective and efficient way? }

To answer \textbf{RQ3} regarding the skills extraction problem, we instantiated our pipeline within the prototype, featuring state-of-the-art machine learning models with hundreds of thousands vacancies stemming from the Dutch-Flemish labour market. In terms of accuracy, the best performing model is BERT Cloud. It also features the fastest execution time, as it leverages on-demand TPU (TensorFlow Processing Units) infrastructure from Google Cloud, resulting in speedups of up to 10x for training and predictions -- from hours/days to minutes. 
That should be taken into account when deploying this prototype in production, given the scalability and performance issues that may arise. 

\subsection{Limitations}
A domain-specific limitation is that our models have been trained on UWV data and encompasses Dutch vacancies and resumes. 
This can bring along cross-border applicability issues as e.g., Flemish and Dutch languages may differ. However, the baseline BERT multilingual models have been used in different contexts (e.g, training in one language and predicting in another) with different languages and outstanding results~\cite{pires2019multilingual}. Another limitation is that the amount of annotated data is still small, and involving more experts in annotating data for training  (both in Dutch and Flemish) is mandatory -- although the results already show high accuracy.

Another concern to be addressed is the data privacy and governance: data should travel back and fort to/from the cloud during the training and prediction processes. This may pose an issue according to GDPR regulations given the sensitivity of the data. As the Werkinzicht project is subject to such regulations, the local solutions may overcome the cloud ones because of privacy and governance, and in spite of performance and scalability.

\section{Conclusions}\label{conc}

This article introduces and explores the design for a DataOps intelligence analytics platform to support a more sustainable labour market. The proposed platform falls into the architecture landscape of what we defined as societal intelligence, that is, the ability of society to become reflective, using big open data analytics to instrument that reflection. 

Our proposed design was partially validated against a real-life case-study from the Dutch borderlands labour acceleration initiative. Moreover, we implemented a prototype for the key tasks of skills extraction from vacancies and resumes, checked against real data from the Dutch employment agency UWV. Results not only show that our appraoch is implementable, but also  that our prototype achieves high results in terms of accuracy and recall, being actually usable implementable in practice.

Meantime, we observed that several technical challenges exist in our proposed design and prototype, that deserve further attention. Conversely, we also highlighted the potential impact behind our proposed design and how it may actually instrument societal intelligence in the Netherlands - an impact that can potentially increase the NL national wealth by up to 5\% (projected). 

In the future we plan to: (a) further refine the social, organizational, and technical requirements behind the proposed design; (b) gather hard data over the data-areas present in our case-study to refine and bring to production our proof-of-concept implementation of the design proposal; (c) Integrate our prototype in the overall pipeline for skills matching against existing ontologies, and (d) Refine such ontologies to reflect the highly dynamic labour market.

\bibliographystyle{IEEEtran}
\bibliography{lab}

\end{document}